\pdfoutput=1

\documentclass[11pt]{article}

\usepackage[]{acl}

\usepackage{times}
\usepackage{latexsym}
\usepackage{multirow}
\usepackage{graphicx}
\usepackage{amsmath}
\usepackage{amssymb}
\usepackage{booktabs}
\usepackage{subcaption}
\usepackage{caption}
\usepackage[T1]{fontenc}

\usepackage[utf8]{inputenc}

\usepackage{microtype}
\usepackage{algorithm}
\usepackage{algorithmic}

\newcommand{\bA}{\mathbf{A}}

\newcommand{\bH}{\mathbf{H}}

\newcommand{\bK}{\mathbf{K}}

\newcommand{\bQ}{\mathbf{Q}}

\newcommand{\bS}{\mathbf{S}}

\newcommand{\bU}{\mathbf{U}}
\newcommand{\bV}{\mathbf{V}}
\newcommand{\bW}{\mathbf{W}}

\newcommand{\bc}{\mathbf{c}}

\newcommand{\bh}{\mathbf{h}}

\newcommand{\bq}{\mathbf{q}}

\newcommand{\bx}{\mathbf{x}}

\newcommand{\ttt}{\texttt}
\newcommand{\ie}{\textit{i.e.}}

\title{Less is More: Learning to Refine Dialogue History \\for Personalized Dialogue Generation}

\author{Hanxun Zhong$^1$, Zhicheng Dou$^{2,}$\thanks{$^*$Corresponding author.} , Yutao Zhu$^3$, Hongjin Qian$^2$, Ji-Rong Wen$^2$ \\
        $^1$School of Information, Renmin University of China, Beijing, China \\ 
        $^2$Gaoling School of Artificial Intelligence, Renmin University of China, Beijing, China \\ 
        $^3$University of Montreal, Quebec, Canada \\
        \texttt{\{hanxun\_zhong,dou,ian\}@ruc.edu.cn} \\
        \texttt{\{yutaozhu94,jirong.wen\}@gmail.com} \\
}

\begin{document}
\maketitle
\begin{abstract}

Personalized dialogue systems explore the problem of generating responses that are consistent with the user's personality, which has raised much attention in recent years. Existing personalized dialogue systems have tried to extract user profiles from dialogue history to guide personalized response generation. Since the dialogue history is usually long and noisy, most existing methods truncate the dialogue history to model the user's personality. Such methods can generate some personalized responses, but a large part of dialogue history is wasted, leading to sub-optimal performance of personalized response generation. In this work, we propose to refine the user dialogue history on a large scale, based on which we can handle more dialogue history and obtain more abundant and accurate persona information. Specifically, we design an MSP model which consists of three personal information refiners and a personalized response generator. With these multi-level refiners, we can sparsely extract the most valuable information (tokens) from the dialogue history and leverage other similar users' data to enhance personalization. Experimental results on two real-world datasets demonstrate the superiority of our model in generating more informative and personalized responses.
\end{abstract}

\section{Introduction}
Recent years have witnessed great progress in building personalized dialogue systems. 
In general, previous work explores building a personalized dialogue system via two pathways: (1) directly modeling user personality from predefined persona descriptions or user attribute~\cite{Huang_2018_kv, Zhang_2018_dog, Song_2019_PCVAE}; and (2) implicitly modeling the user personality from the user's dialogue history~\cite{Li_2016_speaker, ma_dhap_2021}. The latter is considered superior as the dialogue history is easy to obtain and comprises rich personalized information. In this paper, we follow the second pathway that automatically learns implicit user profiles from the user's dialogue history to assist in personalized response generation. 

It is challenging to model user personality directly from the dialogue history. The main reason is that a user's dialogue history may contain massive historical dialogues, which are too heavy to load in the model and likely to be noisy. A straightforward solution is to truncate the dialogue history, as done by existing work~\cite{ma_dhap_2021, qian_imp_2020}. However, as tremendous information has been wasted, the model's performance is also influenced. On the other hand, we observe that the dialogue history from other users may also be helpful in generating a personalized response for the current user. For example, users with the same interest in ``soccer'' may talk about similar things on such a topic. This has been overlooked by existing methods. Intuitively, the problem of ``data explosion'' is even more severe in the latter case (when other similar users' dialogue history is also considered).
To alleviate these problems, we propose using a hierarchical refiner structure to sparsely extract the most valuable query-aware persona information from both the current and other similar users' dialogue history. By this means, more dialogue history can be utilized for learning user personality and improving response generation.

Our model is called \textbf{MSP}, which stands for \textbf{M}odeling and \textbf{S}electing user \textbf{P}ersonality from the dialogue history for generating personalized responses. 
Instead of attending to all dialogue history, MSP refines the most valuable historical information that can well portray the user's personality and guide the response generation. Specifically,  MSP consists of three personal information refiners working at different levels and a personalized response generator.
At \textbf{first}, a user refiner is designed to select a group of users who have similar interests to the current user. By refining dialogue history at the user level, we can obtain similar data to share information with similar users and avoid mutual interference with other users. \textbf{Then}, a topic refiner filters out the current and similar users' dialogue history that has different topics with the current query at the sentence level. \textbf{Next}, we design a token refiner to extract the most valuable query-aware user profiles from the remaining dialogue history at the token level. 
\textbf{Finally}, a personalized response generator combines user profiles and the current query to generate responses. Given that there is no explicit supervisory signal guiding the refiner to extract an exemplary user profile, we design a supplementary sentence matching task and a joint training method. The generator will construct a pseudo-label to guide the refiner's extraction.

Our contributions are three-fold: 

(1) We design an MSP model to tackle the data noise problem. It can efficiently refine user profiles through dialogue history and generate personalized responses.
By this means, our method can capture abundant user profiles while keeping away from noisy data.

(2) We design a refiner structure to extract the query-aware profile at three levels. Similar users' information is taken into account, which can help improve the personality of the response.

(3) We design a joint training method for the refiner and generator. The refiner provides the generator with user profiles to assist in generating responses, while the generator constructs a pseudo-label for the refiner to assist in selecting user profiles.

\begin{figure*}[htbp]
\includegraphics[width=\textwidth]{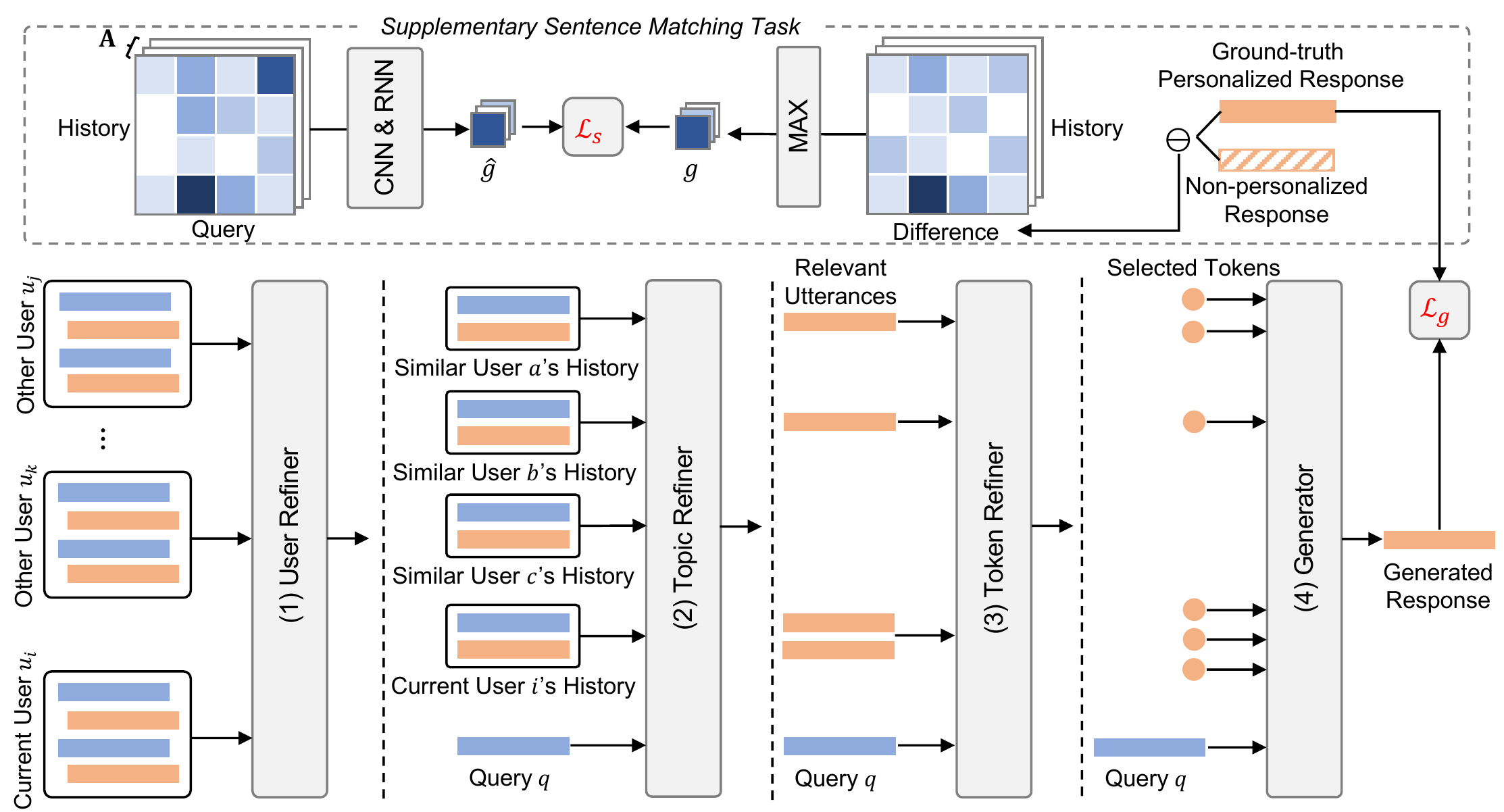}
\caption{The overview structure of the proposed model which consists of four modules: (1) user refiner, (2) topic refiner, (3) token refiner, and (4) generator.}
\label{fig: overview}
\end{figure*}
\section{Related Work}

\paragraph{Personalized Dialogue Generation}
Open-domain dialogue generation has been extensively studied~\cite{koehn_statistical_2003, vinyals_neural_2015,HRED_2016,ReCoSa_2019,Yi_dialoGPT_2019,liu-etal-2020-speaker,ziang_meena_2020,DBLP:journals/ir/ZhuDNW20}. 
Recently, personalized dialogue systems have attached more and more attention. Typical methods include: (1) explicitly using predefined persona descriptions or attributes as users' profile to generate personalized responses~\cite{Huang_2018_kv,Zhang_2018_dog,Alan_2019_PreGan, Song_2019_PCVAE}; (2) using user ID embeddings to enhance personalized dialogue generation~\cite{Li_2016_speaker,yanrui_pwae_2019}; and (3) extracting implicit user profile from users' dialogue history to generate personalized responses~\cite{Rami_CCC_2016,Bak_VHUCM_2019,ma_dhap_2021}. Since manually collecting user profiles is impractical for large-scale datasets and the user ID embeddings perform badly, in this study, we focus on the last group of methods for personalized response generation.

DHAP~\cite{ma_dhap_2021} is the state-of-the-art method in personalized dialogue generation. It uses a transformer-based structure to model the user's dialogue history and extract personal information for response generation. Unfortunately, this model can only handle a limited number of user dialogue histories, wasting a lot of valuable information. Our method has two main differences from DHAP: (1) We propose a refiner structure in our model so that more dialogue history can be handled and the most valuable information can be extracted for improving response generation; (2) With our proposed refiner, we can further incorporate more dialogue history from other users (having similar interests) to facilitate personalized dialogue generation.

\paragraph{Retrieval-guided Natural Language Generation} Retrieval-based methods can collect relevant information for language generation~\cite{Liu_hybrid_2019}. It has been widely applied in many tasks such as text style transfer~\cite{Li_drg_2018} and dialogue generation~\cite{Wu_prototype_2019, Cai_m2g_2019}. The idea of using a retrieval system to get useful information inspires our study. We use a refiner to automatically extract personal information from dialogue history and guide the personalized generation.

\section{Methodology}
In this section, we first formalize our problem and provide an overview of our proposed MSP model. Then, we describe the details of each component and the model optimization.
\subsection{Problem Statement and Overview}
Considering a set of users $\mathcal{U}=\{u_1,\cdots,u_l\}$, for any user $u_i$, we have their dialogue history with others $U^i=\{(q_1^i,r_1^i),\cdots,(q_t^i,r_t^i)\}$, where $q_j^i$ is a \textit{query} issued by others, while $r_j^i$ is the corresponding \textit{response} given by $u_i$.\footnote{Here we use the term ``query'' to denote the utterance given by others. Generally, the query can be either one utterance in single-turn dialogues, or several history utterances in multi-turn dialogues.} Our target is to generate a personalized response $r^i$ for the user $u_i$ to reply a new query $q$. As we introduced earlier, the personalized information can be obtained from the dialogue history $U^i$ of the user $u_i$ and other dialogue history $U^j$ from similar users $u_j(j\neq i)$.

The overview of our MSP model is shown in Figure~\ref{fig: overview}. MSP consists of three personal information refiners working on different levels and a personalized response generator. Specifically, the first refiner is working at the user level. By comparing the dialogue history of the current user $u_i$ with others, MSP can select a group of users having similar interests with $u_i$. 
After obtaining a group of similar users, we further refine their dialogue history according to the relevance to the current query's topic. Moreover, we add the last refiner to extract several tokens so that the most fine-grained personal information can be extracted from the relevant utterances. Finally, the query and the extracted tokens are fed into the generator and construct a personalized response.

\subsection{User Refiner}
The dialogue history of users with similar interests may share much personal information. Therefore, our first target is to select a group of users with similar interests to the current user. We design a user refiner to achieve this.
Since the users' interest is usually contained in their dialogues with others, we consider both the queries and responses in the dialogue history to select similar users.
Specifically, for the user $u_i$'s dialogue history ${U}^i$, we apply a pre-trained BERT~\cite{Jacob_bert_2019} and represent them by the embedding of \ttt{[CLS]} token:
\begin{align*}
 {\bU}_q^i &= \sum_{j=1}^{t} \operatorname{BERT}(q_j^i), \quad
 {\bU}_r^i &= \sum_{j=1}^{t} \operatorname{BERT}(r_j^i).
\end{align*}
Then, we can select $k_u$ users that have similar interest with the current user $u_i$ as:
\begin{align}
    {u}^\text{sim} &= \operatorname{TopK}(\bU^i\cdot\bU^j,k_u), \\
    \bU^i &= [{\bU}_q^i; {\bU}_r^i],
\end{align}
where $\operatorname{TopK}(\cdot,\cdot)$ is the top-$k$ selection operation. 

After the user refiner, we can obtain the dialogue history of the similar users $\{u_j\}_{j=1}^{k_u}$. It is worth noting that, since the number of users is large in the datasets, we choose to use the dot-product to compute the similarity of the users so that the whole process can be implemented by dense retrieval libraries, such as Faiss~\cite{JDH_faiss_17}, which is very efficient.

\subsection{Topic Refiner}
The users' dialogue history often contains many dialogues with others. These dialogues have various topics, which may be irrelevant to the current query. Therefore, we propose a topic refiner to select relevant dialogue history for personalized response generation.
Specifically, we use a topic classifier to compute the topic distribution of the current query $q$ and the queries $q_j^i$ in the history dialogues:
\begin{align}
    t &= \text{MLP}(\operatorname{mean}(\operatorname{BERT}(q)), \\
    t_j^i &= \text{MLP}(\operatorname{mean}(\operatorname{BERT}(q_j^i)),
\end{align}
where $t,t_j^i \in \mathbb{R}^{d^t \times 1}$, and $d^t$ is the number of topic. Then, we filter out the dialogue history $<q_j^i,r_j^i>$ that has different topics with the current query, \ie, ${\max}(t_j^i) \neq {\max}(t)$.

In the topic refining process, we compare the queries in the history dialogues with the current query to filter out topic-irrelevant dialogues. This process can further reduce the noise and make our model more lightweight. Both the dialogue history of the current user and that of the similar users (obtained in the former step) are refined. In the next step, we will use the \textbf{responses} in these selected history dialogues and extract the most valuable tokens for personalized response generation.

\subsection{Token Refiner}
After the previous two refiners, we obtain a collection of historical responses. Though we can directly add them into the generation process, our preliminary experiments indicate that they perform poorly. A major reason is the noisy quality of the responses. Indeed, existing studies~\cite{DBLP:conf/aaai/XingWWLHZM17,DBLP:journals/ir/ZhuDNW20} have demonstrated the effectiveness of using informative tokens to improve the response generation. Inspired by these studies, we further devise a token refiner to extract the most fine-grained information (tokens) from the historical responses.
Specifically, we compute an attention map $\bA$ between the query $q$ and the historical responses $r^{\text{sim}}$ and $r^\text{cur}$ (they are from the similar users and the current user respectively) as:
\begin{align}
 \bA &= \operatorname{softmax}\left(\frac {\bQ \bK^{T}}{\sqrt{d}}\right), \\
 \bQ &= \text{TRM}_\text{enc}(q) \cdot \bW_Q, \\
 \bK &= \text{TRM}_\text{enc}(r) \cdot \bW_K, 
\end{align}
where $\text{TRM}_\text{enc}(\cdot)$ is a transformer encoder. $r$ refers to $r^{\text{sim}}$ or $r^\text{cur}$, and correspondingly, $\bA$ refers to the similar user matching map $\bA^\text{sim}$ and current user matching map $\bA^\text{cur}$. $\bW_Q, \bW_K\in \mathbb{R}^{d \times d}$ are parameters, and $d$ is the dimension of the hidden state. After obtaining the attention matching map $\bA$, we select tokens to form the similar users' profile and current user's profile according to each token's attention weight:
\begin{align}
    \bc^{sim} &= \operatorname{TopK}(\operatorname{Max}(\bA^\text{sim}), k_p), \label{ccim}\\
    \bc^{per} &= \operatorname{TopK}(\operatorname{Max}(\bA^\text{cur}), k_p), \label{cper}
\end{align}
where $k_p$ is a hyper-parameter to control the number of profile tokens. Its influence will be investigated in our experiments.

\subsection{Generator}
We use a transformer decoder as to generate a personalized response by using the similar users' profile $\bc^\text{sim}$, current user's profile $\bc^\text{cur}$, and query information $\bq$ as input. The decoding process can be defined as:
\begin{align}
    \hat{y} &= \text{TRM}_\text{dec}(\bx), \\
    \bx &= [\bc^\text{sim}; \bc^\text{per}; \bq], 
\end{align}
where [;] is the concatenation operation and $\hat{y}$ is the word generation probability. 

\subsection{Training and Optimization}
The generator is optimized by maximizing the generation probability of the ground-truth $y$:
\begin{align}
    \mathcal{L}^{\text{g}} = -y\log\hat{y}. \label{lg}
\end{align}


In practice, we find that the token refiner is hard to train. We speculate the reason is a missing of direct supervision signals. In this case, it is difficult to tell whether the training errors stem from the generation process or the refining process. To tackle this problem, we propose a supplementary sentence matching task to assist the token selection. 

\paragraph{Supplementary Sentence Matching Task}
The core idea of this task is to train the token refiner directly by introducing supervision signals so that it can automatically mine valuable tokens.
Specifically, we design a sentence matching task to match the query with the dialogue history. The task's target is to find the history sentences that help generate personalized responses. We consider using the query-history cross attention weight $\bA$ to generate a matching representation and then use this representation to finish the task. In this way, once the matching task has been well-finished, we can use attention map $\bA$ to identify the most informative and valuable parts of a history sentence that are helpful to generate a personalized response.

To achieve our idea, we design a matching process. First, we calculate the matching representations $\bH$ by the cross-attention map $\bA$ and then apply a CNN with a max-pooling operation to aggregate token information:
\begin{align}
    \bS &= \operatorname{Maxpool}(\text{CNN}(\bH)) \\
    \bH &= \bA \cdot \bV, \\
    \bV &= \text{TRM}_\text{enc}(r) \cdot \bW_V.
\end{align}
Next, we flatten $\bS$ and applies a LSTM to aggregate the sentence information:
\begin{align}
   \bh = \text{LSTM}(\operatorname{Flatten}(\bS)). 
\end{align}
Finally, we use the sentence matching vector $\bh$ to compute the matching score:
\begin{align}
    \hat{g} = \operatorname{Sigmoid}(\text{MLP}(\bh)).
\end{align}
For guiding the sentence matching task, we design a pseudo-label $g$ to measure the matching goodness of each history sentence. We expect the history with more persona profile information can achieve a higher score. Therefore, we use the difference between the personalized ground-truth $y$ and a non-personalized generated response $\hat{y}^{\prime}$ to measure the persona profile and create the pseudo-label: 
\begin{align}
 g &= 
 \begin{cases}
 1, & g_{\rm soft} \geq \alpha \\
 0, & g_{\rm soft} \textless \alpha,
 \end{cases}\label{psesudo}
 \\
 g_{\rm soft} &= \operatorname{Sum}(\operatorname{Max}((y - \hat{y}^{\prime}) \cdot r)) / d_\text{y}, 
\end{align}
where $d_\text{y}$ is the length of $y$, and $\alpha$ is a threshold. Finally, we minimize the binary cross entropy loss between $g$ and $\hat{g}$:
\begin{align}
\mathcal{L}^{\text{s}} &= g\log\hat{g} + (1-g)\log(1-\hat{g}). \label{ls}   
\end{align}

\paragraph{Joint Training} To facilitate the learning with the above gradient approximation approach, we design a joint training process to train the refiner and generator in turn. Specifically, in each training iteration, we first sample a batch of query $q$, response $y$, similar and current users' dialogue history $r^\text{sim}$ and $r^\text{cur}$ from dataset $D$. Then, we generate a non-personalized response $\hat{y}^{\prime}$ and create the pseudo-label $g$ (Eq.~\ref{psesudo}) through a non-personalized generator. This pseudo-label is used to train the token refiner by optimizing the loss $\mathcal{L}^{\text{s}}$ (Eq.~\ref{ls}). Further, we sample another batch $D^p$ from $D$. After extracting similar user profile $\bc^\text{sim}$ and current user profile $\bc^\text{cur}$ (Eq.~\ref{ccim}, Eq.~\ref{cper}), we generate the personalized response $\hat{y}$ and update the generator by optimizing the loss $\mathcal{L}^{\text{g}}$ (Eq.~\ref{lg}). To avoid misleading the generation (by poor profile) at the beginning of the training process, we pre-train the refiner for $N_f$ steps before extracting the profile for the generator. The detailed training process is summarized in Algorithm~\ref{alg1}.      
\begin{algorithm}[t]
    \small
	\caption{Joint Training Process} 
	\label{alg1} 
	\begin{algorithmic}[1]
		\REQUIRE $M$ dialogue triplets:$D={\{\langle q_i, y_i, r_i \rangle \}^M_{i=1}}$
		\ENSURE A personalized dialogue model
		\STATE Init the refiner and generator module
		\WHILE{not converge}
		    \STATE Sample $n_s$ dialogue triplets $D^q = {\{ q_i, y_i, r_i\}^{n_s}_{i=1}}$
		    \STATE Get $\hat{Y}^q = {\{ \hat{y}^{\prime}_i\}^{n_s}_{i=1}}$ on $D^q$ from $p_g\left(\hat{y}^{\prime}|q\right)$
		    \STATE Get pseudo-label $g$ on $D^q$
		    \STATE Train refiner by optimizing $\mathcal{L}^{s}$ on $D^q$
		    \IF{Current Step \textgreater $N_f$}
    		    \STATE Sample $n_d$ dialogue triplets $D^p={\{ q_i, y_i,r_i\}^{n_d}_{i=1}}$
    		    \STATE Extract $C^p=\{c_i\}^{n_d}_{i=1}$ on $D^p$ from $p_s\left(c|q,r\right)$ 
    		    \STATE Train generator by optimizing $\mathcal{L}^{g}$ on $D^p \cup C^p$
		    \ENDIF
		\ENDWHILE
	\end{algorithmic}
\end{algorithm}

\begin{table}[t]
\centering
\small
\caption{Statistics of Reddit and Weibo datasets. }
\begin{tabular}{lrr}
\toprule
& \textbf{Reddit} & \textbf{Weibo} \\ 
\midrule
\# Users & 78,031 & 46,973 \\
Avg. history length & 72.4 & 30.8 \\
Avg. \# words of query & 19.8 & 22.9\\
Avg. \# words of response & 9.1 & 9.6\\
\# Training samples & 5,734,129 & 1,495,149 \\
\# Validation samples & 10,000 & 10,000 \\
\# Testing samples & 10,000 & 10,000 \\
\bottomrule
\end{tabular}
\label{tab: dataset}
\end{table}

\begin{table}[t]
\caption{Criteria of human annotation.}
\small
\centering
\label{tab: cri}
\begin{tabular}{l}
\toprule
\textbf{Readability} \\
3: Fluent and easy to read \\ 
2: Grammatically formed \\  
1: Not a complete sentence or hard to read \\  
\midrule
\textbf{Informativeness} \\ 
3: Have clear and specific meaning \\
2: Contain few informative words \\ 
1: Meaningfulness sentence  \\
\midrule
\textbf{Personalization} \\
1: Reflect personal information contained in user history \\
0: Does not resemble any user history \\ 
\bottomrule
\end{tabular}
\end{table}

\section{Experiments}
\subsection{Datasets}
To evaluate our model's performance, we conduct experiments on a Chinese Weibo dataset~\cite{pchatbot_sigir} and an English Reddit dataset.\footnote{Weibo: \url{https://www.weibo.com/}; Reddit: \url{https://www.reddit.com/}} Both are collected from open-domain social media platforms. On these platforms, users can post various topics, and other users can respond to them. We compare user-id and timestamps to associate the query with its corresponding response and the current user's dialogue history. As a result, each training sample contains a query, a response, and a sequence of dialogue history. Finally, the dataset is divided into training, validation, and test sets in chronological order. The statistics of the datasets are provided in Table~\ref{tab: dataset}.
\subsection{Baseline Methods}
We compare our proposed model with eight highly correlated and strong baselines. They can be categorized into four groups:

\paragraph{Non-personalized Methods} (1) {Seq2Seq-Attention}~\cite{Ilya_S2S_2014} is a vanilla sequence-to-sequence model with attention mechanism ~\cite{ Luong_atten_2015}.
(2) {MMI}~\cite{Li_mmi_2016} is based on Seq2seq and use maximum mutual information as an extra loss to improve diversity. (3) {DialoGPT}~\cite{Yi_dialoGPT_2019} is a variant of GPT-2~\cite{radford_gpt2_2019} designed for dialogue generation.

\paragraph{Predefined Profile-based Methods} Since there is no persona description in the datasets, we test these methods by using the user's dialogue history as a simulation of predefined persona profiles. (4) {GPMN}~\cite{Zhang_2018_dog} enhances the Seq2seq model with a memory module, which encodes and stores the persona profile as memory representations. (5) {PerCVAE}~\cite{Zhao_2017_CVAE} encodes predefined personalized sentences as a conditional representation and uses CVAE to generate a personalized response.

\paragraph{User ID-based Methods} (6) {Speaker} ~\cite{Li_2016_speaker} is based on seq2seq while using user-ID embedding as user representation to facilitate the response generation. 
(7) {Persona WAE}~\cite{yanrui_pwae_2019} uses WAE (Wasserstein autoencoder) for response generation. It maps user-ID embeddings to a personalized Gaussian mixture distribution and then samples the personalized vector to guide the response generation.

\paragraph{User Dialogue History-based Methods} 
(8) {DHAP}~\cite{ma_dhap_2021} uses history memory to store and construct the dynamic query-aware user profile from dialogue history and then uses a personalized decoder to generate a response. Since this model also learns the user profile directly from the dialogue history, it is the most relevant baseline of our method.

\begin{table*}[t!]
\caption{Metric-based evaluation result on Weibo and Reddit dataset. The best results are in \textbf{bold}. ``$\dagger$'' indicates that our model achieves significant improvement in t-test with $p$-value $<$ 0.05.}\label{tab: auto}
\small
\centering
\setlength{\tabcolsep}{1.5mm}{
\begin{tabular}{llrrrrrrrrrr}
\toprule
&  & \multicolumn{3}{c}{Overlap-based Metric} & \multicolumn{2}{c}{Diversity} & \multicolumn{3}{c}{Embedding Metric} & \multicolumn{2}{c}{Persona Metric} \\  
\cmidrule(lr){3-5}\cmidrule(lr){6-7}\cmidrule(lr){8-10}\cmidrule(lr){11-12} 
& & {BLEU-1} & {BLEU-2} & {ROUGE-L} & {Dist-1} &  {Dist-2} & {Average} &  {Extrema} &  {Greedy} &  {P-F1} &  {P-Cover}\\ 
\midrule
\multirow{9}{*}{\rotatebox[origin=c]{90}{Weibo}} & Seq2Seq & 3.330$^\dagger$ & 0.294$^\dagger$ & 8.985$^\dagger$ & 0.930$^\dagger$ & 2.180$^\dagger$ & 0.321$^\dagger$ & 0.266$^\dagger$ & 0.254$^\dagger$ & 0.154$^\dagger$ & 0.041$^\dagger$\\
 & MMI & 3.631$^\dagger$ & 0.095$^\dagger$ & 5.264$^\dagger$ & 10.710$^\dagger$ & 43.458$^\dagger$ & 0.477$^\dagger$ & 0.696$^\dagger$  & 0.305$^\dagger$ & 0.325$^\dagger$ & 0.054$^\dagger$\\
 & DialoGPT & 6.068$^\dagger$ & 0.741$^\dagger$ & 8.459$^\dagger$  & 15.322$^\dagger$ & 55.536$^\dagger$ &0.557$^\dagger$ & 0.793$^\dagger$ & 0.324$^\dagger$ & 0.522$^\dagger$ & 0.061$^\dagger$\\
 & GPMN & 4.899$^\dagger$ & 0.696$^\dagger$ & 7.785$^\dagger$ & 11.724$^\dagger$&  32.730$^\dagger$ & 0.353$^\dagger$ & 0.391$^\dagger$  & 0.301$^\dagger$ & 0.542$^\dagger$ & 0.084$^\dagger$ \\
 & PerCVAE & 5.114$^\dagger$ & 0.299$^\dagger$& 7.380$^\dagger$ & 14.098$^\dagger$ & 49.733$^\dagger$ & 0.469$^\dagger$ & 0.657$^\dagger$ & 0.299$^\dagger$ & 0.903$^\dagger$ & 0.086$^\dagger$\\
 & Speaker & 4.994$^\dagger$ & 0.113$^\dagger$ & 7.868$^\dagger$ & 6.035$^\dagger$ & 19.007$^\dagger$ & 0.492$^\dagger$ & 0.712$^\dagger$ & 0.311$^\dagger$ & 0.225$^\dagger$ & 0.082$^\dagger$\\
 & PersonaWAE & 3.510$^\dagger$ & 0.155$^\dagger$ & 10.546$^\dagger$ & 2.551$^\dagger$ & 19.743$^\dagger$ & {0.563}$^\dagger$ & {0.757}$^\dagger$ & 0.307$^\dagger$ & 1.740$^\dagger$ & 0.103$^\dagger$\\
 & DHAP & {9.324}$^\dagger$ & {0.894}$^\dagger$& {14.122}$^\dagger$ & {15.175}$^\dagger$ & {58.806}$^\dagger$  & 0.523$^\dagger$ & 0.747$^\dagger$ & {0.313}$^\dagger$ & {1.791}$^\dagger$ & {0.144}$^\dagger$ \\
 & {MSP} (Ours) & \textbf{11.875} & \textbf{5.108}& \textbf{15.563} & \textbf{24.203}  & \textbf{73.196} & \textbf{0.605} & \textbf{0.883} &  \textbf{0.331} & \textbf{2.170} & \textbf{0.297}\\ 
\midrule
\multirow{10}{*}{\rotatebox[origin=c]{90}{Reddit}} & Seq2Seq & 1.820$^\dagger$ & 0.023$^\dagger$ & 4.069$^\dagger$  & 5.203$^\dagger$ & 19.485$^\dagger$ & 0.545$^\dagger$ & 0.554$^\dagger$ &  0.470$^\dagger$ & 0.051$^\dagger$ & 0.029$^\dagger$\\
 & MMI  & 2.065$^\dagger$& 0.011$^\dagger$ & 3.784$^\dagger$ & 5.914$^\dagger$ & 31.093$^\dagger$ & 0.543$^\dagger$ & 0.607$^\dagger$ & 0.454$^\dagger$ & 0.085$^\dagger$ & 0.038$^\dagger$\\
 & DialoGPT & 4.735$^\dagger$ & 0.397$^\dagger$ & 8.943$^\dagger$ & 6.353$^\dagger$ & 29.106$^\dagger$ & 0.604$^\dagger$  & 0.733$^\dagger$ & 0.448$^\dagger$ & 0.137$^\dagger$ & 0.040$^\dagger$\\
 & GPMN & 2.686$^\dagger$ & 0.376$^\dagger$ & 4.776$^\dagger$ & 12.325$^\dagger$ & 35.762$^\dagger$ & 0.406$^\dagger$  & 0.331$^\dagger$ & 0.358$^\dagger$ & 0.189$^\dagger$ & 0.037$^\dagger$\\
 & PerCVAE & 5.933$^\dagger$ & 0.576$^\dagger$ & 8.112$^\dagger$ & 9.631$^\dagger$ & 40.213$^\dagger$  & 0.637$^\dagger$ & 0.649$^\dagger$ & 0.499$^\dagger$ & 0.212$^\dagger$ & 0.040$^\dagger$\\
 & Speaker & 2.642$^\dagger$ & 0.054$^\dagger$ & 4.469$^\dagger$ & 8.951$^\dagger$ & 34.187$^\dagger$ & 0.538$^\dagger$ & 0.606$^\dagger$ & 0.457$^\dagger$ & 0.115$^\dagger$ & 0.031$^\dagger$\\
 & PersonaWAE & 2.637$^\dagger$ & 0.113$^\dagger$ & 8.199$^\dagger$ & 1.758$^\dagger$ & 25.915$^\dagger$ & 0.629$^\dagger$  & 0.685$^\dagger$ & 0.442$^\dagger$ & 0.206$^\dagger$ & 0.032$^\dagger$\\
 & DHAP & 6.858$^\dagger$ & 0.737$^\dagger$ & 11.720$^\dagger$ & 18.707$^\dagger$ & 66.932 & 0.709  & 0.721$^\dagger$ & 0.539 & 0.227$^\dagger$ & 0.111$^\dagger$\\
 & {MSP} (Ours) & \textbf{7.174} & \textbf{0.883} & \textbf{12.171} & \textbf{21.247} & \textbf{68.897} & \textbf{0.716} & \textbf{0.764} & \textbf{0.545} & \textbf{0.276} & \textbf{0.137} \\ 
\bottomrule
\end{tabular}%
}
\end{table*}

\subsection{Implementation Details}
\label{sec:implement}
We experiment with multiple sets of hyperparameters to select the best model, and the final hyperparameters are listed as follows: The dimensions of the embeddings and Transformer hidden units are 768. The number of heads in the Transformer is 12. The number of layers in the Transformer is set as 2 and 12 respectively in the query encoder and decoder. The topic number is 15, and the similar user number is set as 10. The selected profile token number is 200 for the Weibo dataset and 30 for the Reddit dataset. The batch size is 128. Following~\cite{Li_topp_2020}, we adopt nucleus sampling as our decoding strategy. We use the Adam~\cite{DBLP:journals/corr/KingmaB14} optimizer for training the refiners and AdamW~\cite{DBLP:conf/iclr/LoshchilovH19} with a warm-up method for the generator. Our code is publicly available.\footnote{\url{https://github.com/bangbangbang12315/MSP/tree/release}.}

\subsection{Evaluation}
\paragraph{Metric-based Evaluation} We first evaluate all methods by several metrics with respect to different aspects. (1) BLEU-1/2~\cite{Salim_bleu_2002} and ROUGE-L~\cite{Donia_rouge_2004} are typical word overlap-based metrics for measuring the similarity between the generated response and the ground-truth.\footnote{The results of BLEU-3/4 and ROUGE-1/2 are provided in Appendix~\ref{sec:extra}.}
(2) Distinct-1/2~\cite{Jiwei_distinct_2016} consider the number of uni- or bi-grams in the generated response, which is commonly used for evaluating the diversity. (3) The embedding-based metrics (\ie, average, extrema, and greedy)~\cite{yanrui_pwae_2019} are introduced to measure the semantic similarity between the generated response and the ground-truth one. (4) As a personalized dialogue model, following previous studies~\cite{ma_dhap_2021}, two tailored metrics are adopted to measure how much information is included in the dialogue history that can be reflected in the response. Persona-F1 (P-F1)~\cite{Lian_PF1_2019} calculates the F1 value to measure the uni-grams co-occurring in both the generated response and the dialogue history. Persona Coverage (P-Cover)~\cite{Song_2019_PCVAE} calculates the IDF-weighted word overlap between the generated response and the golden one so that the importance of different words can be taken into account.

\paragraph{Human Annotation}
Due to the variability of human language, a response that differs from the ground-truth may also be appropriate. Following previous studies~\cite{yanrui_pwae_2019}, we conduct a human evaluation of all methods. Concretely, we sample 100 (query, response, user dialogue history) triplets and hire three well-educated annotators to score the responses generated by different models. Three aspects, \ie, readability, informativeness, and personalization, are considered. The first two factors are scored on a scale of [1, 3] for their quality, while the third is assessed on a scale of [0, 1], indicating whether the response can accurately reflect the user's personality. The detailed scoring criteria are shown in Table~\ref{tab: cri}.

\subsection{Experimental Results}
\paragraph{Metric-based Evaluation} Table~\ref{tab: auto} shows all models' performance under different metrics. On both datasets, it is clear to see that our MSP model outperforms baselines on all metrics. The improvement is statistically significant (t-test with $p$-value $<$ 0.05). These findings indicate that our model is capable of generating more fluent, diverse, and personalized responses than all baselines. In particular, we can observe:
(1) MSP achieves better performance on overlap-based metrics. This suggests that our model can provide responses that are more similar to the ground-truth with the help of the selected tokens. 
(2) For diversity metrics, the higher distinct values show that our generated responses are more diverse. Additionally, predefined profile-based methods and user dialogue history-based methods outperform others. This shows that incorporating external information can aid in the generation of more informative responses.
(3) In addition to generating more overlapped words with the ground-truth response, the improvements in embedding metrics reflect that our model generates more semantically relevant responses.
(4) Finally, the increase in personalized metrics implies that our approach can incorporate more user-specific information into the generation. Furthermore, the significant improvement over DHAP demonstrates that our model can extract more meaningful personalized information from the user dialogue history.

\paragraph{Human Annotation}
The result of human annotation on the Weibo dataset is shown in Table~\ref{tab: human}. The Fleiss Kappa is around 0.62, indicating a substantial agreement achieved by three annotators. In general, the results of human annotation are consistent with those of the metric-based evaluation. Both of them demonstrate our model's superiority in generating more fluent, informative, and personalized responses. Compared to non-personalized methods, user id-based methods can enhance personalization at the expense of readability. User dialogue history-based methods (\ie, DHAP and MSP) can largely improve the personalization of the response while retaining a high level of readability and informativeness. We attribute this to the abundant personal information contained in the user dialogue history. 
\begin{table}[t!]
\centering
\small
\caption{The result of human evaluation on Weibo dataset. ``$\dagger$'' indicates that our model achieves significant improvement in t-test with $p$-value $<$ 0.05.}
\label{tab: human}
\setlength{\tabcolsep}{1.3mm}{
\begin{tabular}{lrrr}
\toprule
Model & Readability & Informativeness & Personality \\ \midrule
Seq2Seq & 1.76$^\dagger$ & 1.37$^\dagger$ & 0.11$^\dagger$  \\
MMI & 1.96$^\dagger$ & 1.88$^\dagger$ & 0.19$^\dagger$ \\
DialoGPT  & {2.33} &  {2.10}$^\dagger$  & 0.32$^\dagger$  \\
GPMN  & 2.01$^\dagger$ &  {2.16}$^\dagger$  & 0.35$^\dagger$  \\
PerCVAE  & 2.10$^\dagger$ &  {2.01}$^\dagger$  & 0.39$^\dagger$  \\
Speaker & 1.89$^\dagger$ &  1.44$^\dagger$  & 0.24$^\dagger$  \\
PersonaWAE & 1.81$^\dagger$ &  2.01$^\dagger$  & 0.32$^\dagger$  \\
DHAP & 2.29$^\dagger$ &  2.19$^\dagger$  &  {0.55}$^\dagger$ \\ 
{MSP (Ours)}  &  \textbf{2.37} & \textbf{2.39}  & \textbf{0.67} \\ \midrule
Ground-Truth &  2.71 &  2.66  &  0.76 \\ \bottomrule
\end{tabular}%
}
\end{table}

\begin{table}[t!]
\centering
\small
\caption{The results of ablation experiments on Weibo dataset.}
\label{tab: ablation}
\setlength{\tabcolsep}{1.5mm}{
\begin{tabular}{lrrr}
\toprule
Models & BLEU-1 & BLEU-2 & P-Cover \\ \midrule
MSP (Full) & \textbf{11.875} &  \textbf{5.108} & \textbf{0.297} \\
\quad\textit{w/o} User Refiner & 6.093 &  0.757 & 0.151 \\
\quad\textit{w/o} Topic Refiner & 6.163 & 0.839 & 0.178 \\
\quad\textit{w/o} Token Refiner & 4.213 &  0.609 & 0.116 \\
\quad\textit{w/o} Current U's Profile & 9.365 & 3.146 & 0.238 \\ 
\quad\textit{w/o} Similar Us' Profile & 6.413 & 0.871 &0.245 \\ 
\quad\textit{w/o} Joint Training & 6.070 & 0.749  & 0.130 \\ 
\bottomrule
\end{tabular}%
}
\vspace{-5pt}
\end{table}

\section{Further Analysis}
We further conduct a series of analyses to elaborate our model. All analyses here are based on the result of the Weibo dataset, while similar results can be observed on the Reddit dataset.

\paragraph{Ablation Study} To investigate the impact of different modules in MSP, we conduct an ablation study by removing or using different strategies in each module. 

We first study the influence of the refiners at three levels: (1) We remove the user refiner and train our model using randomly sampled users. We can see the performance of all metrics drops. This illustrates that our MSP model can select users that share the same interests as the current user and thereby improving response quality. (2) We remove the topic refiner and supply the token refiner with full dialogue history. The performance degradation demonstrates that various topics in dialogue history introduce lots of noise, misleading the token refiner on extracting valuable tokens, thus impairing the personalized response generation. (3) We eliminate the token refiner and feed all dialogue history sentences directly into the generator.\footnote{Due to the length limitation of GPT-2, history with more than 512 tokens will be truncated.} The decline in performance implies the effectiveness and necessity of token selection. It is worth noting that, as compared to using the complete history, our selection strategy can reduce training time by 41.6\%, considerably increasing efficiency. All of the aforementioned experimental results suggest that MSP's advantage stems from high-quality personalized information extraction rather than simply introducing additional information.

We then explore the impact of personalized information from two sources, \ie, the current user's profile and the similar users' profile. Removing either of them results in decreased performance. This exemplifies their usefulness. Specifically, compared with similar users' profiles, eliminating the current user's profile will hurt the personalization effect heavily. This result shows that, for personalization, the current user's profile is more essential than that of similar users, which is quite intuitive. However, the similar users' profile has a significant effect on BLEU-1/2, implying that such a profile can provide abundant information for response generation. Consequently, integrating both types of profiles significantly improves response generation.

Finally, we conduct an experiment to validate our proposed joint training for the token refiner. The declining performance indicates that the token refiner is unable to extract useful information in the absence of additional supervision signals. Indeed, when the sentence matching task is removed, the token refiner extracts tokens that are relevant to the current query, which is less useful for generating a personalized response.

\begin{figure}[t!]
    \centering
    \includegraphics[width=\linewidth]{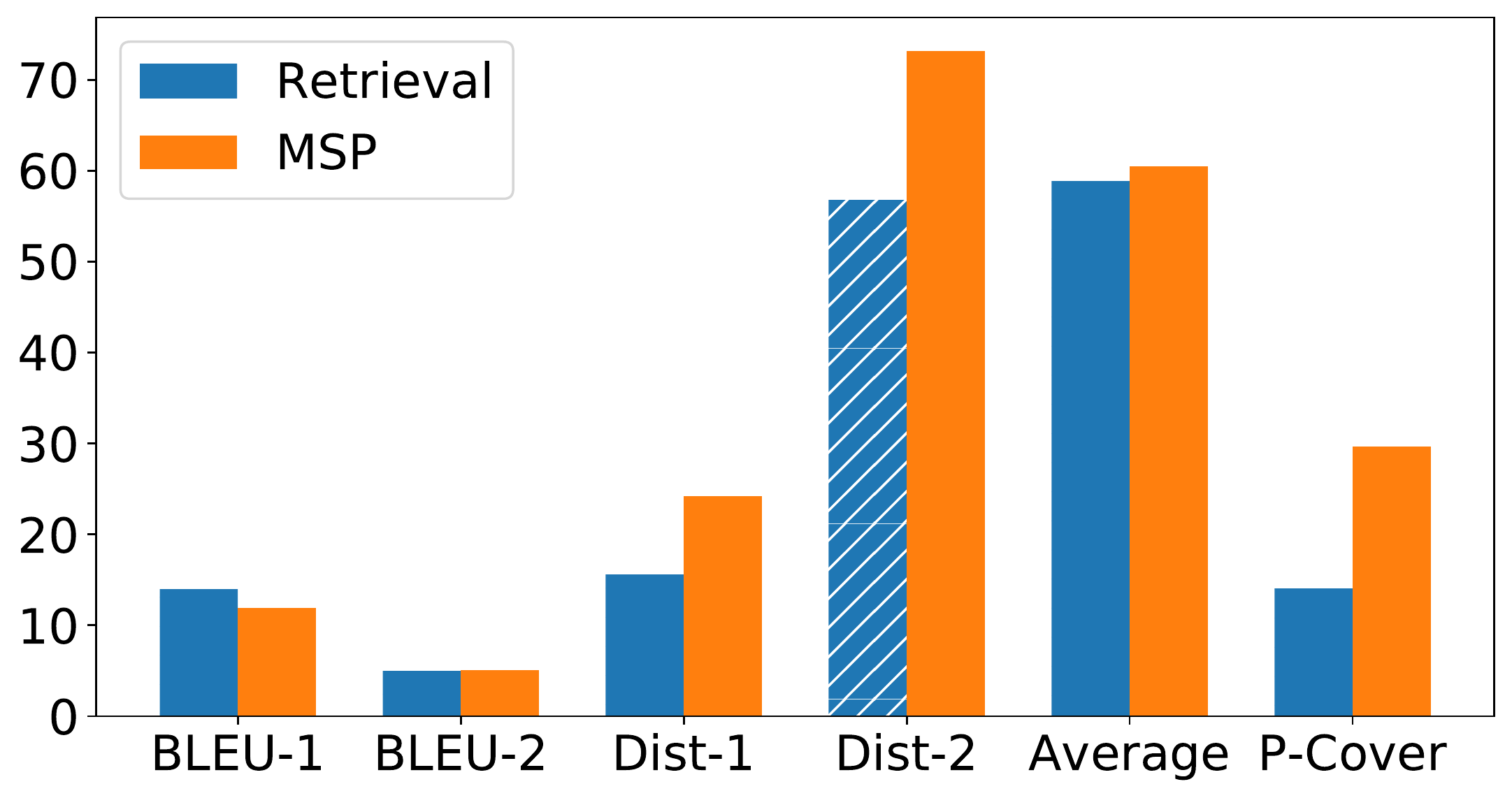}
    \caption[Caption for LOF]{Comparison with the retrieval-based model on the Weibo dataset.}
    \label{fig: retrieval}
\end{figure}

\begin{figure}[t!]
    \centering
    \includegraphics[width=\linewidth]{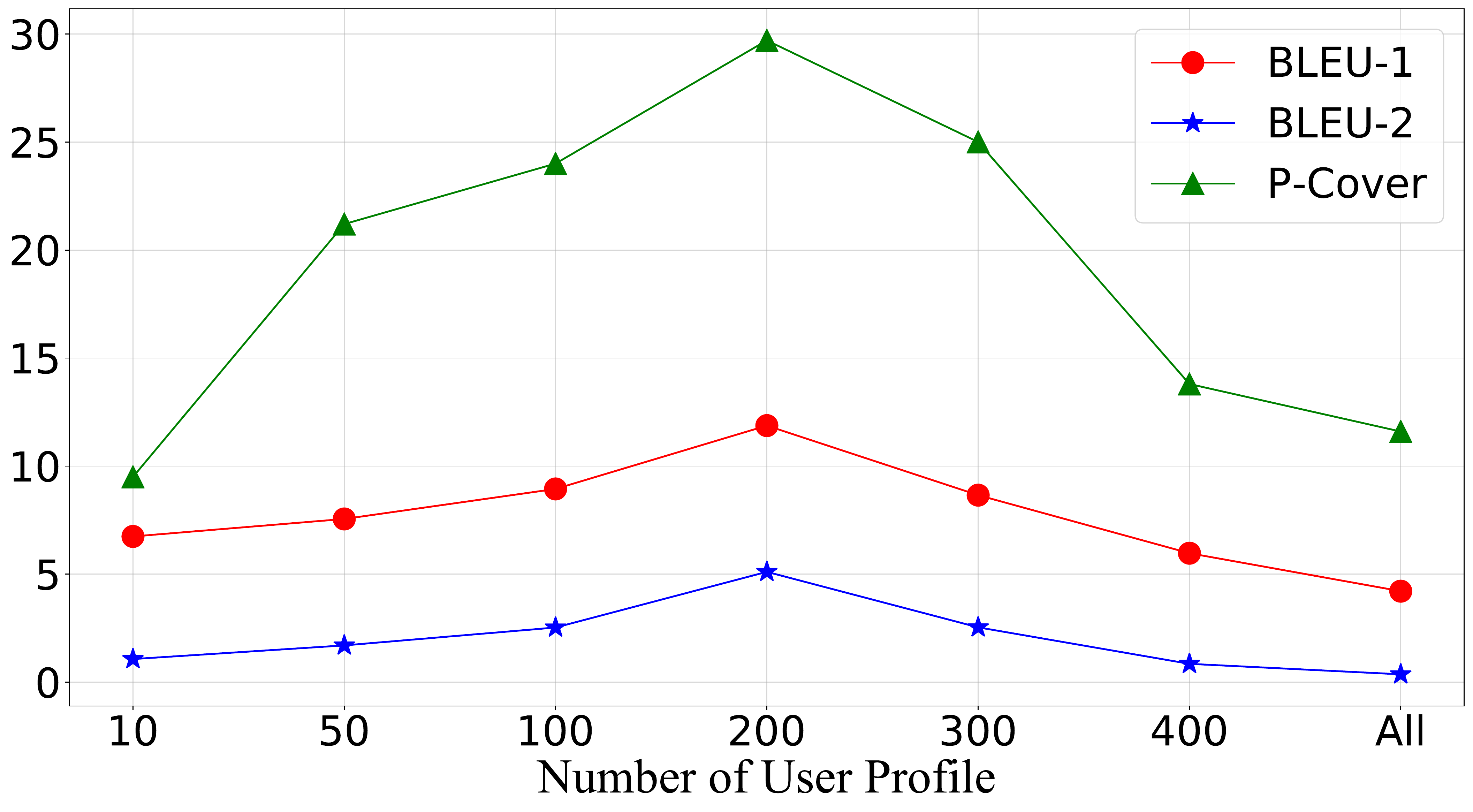}
    \caption[Caption for LOF]{Experiments with the different number of user profiles on the Weibo dataset. To keep the dimension consistent, P-Cover is multiplied by a factor of 100.}
    \label{fig:his}    
\end{figure}
\paragraph{Influence of Selection Mechanism}
To validate the effectiveness of our proposed selection mechanism, we replace the refiner with a traditional retrieval method (\ie, BM25~\cite{bruce_bm25_1994}). Specifically, we use the query to retrieve 15 relevant responses and feed them into our model for training. The experimental results are shown in Figure~\ref{fig: retrieval}. 
We can observe that the retrieval strategy achieves comparable performance with our model on word-overlap and embedding-based metrics. This suggests that the relevant dialogue history for the query can provide valuable information for response generation. However, the retrieval strategy performs poorly on diversity and personalization metrics. This demonstrates that, without careful selection, the retrieved information is too generic and thus less helpful for personalized response generation.

\paragraph{Influence of Personalized Tokens Amount} In MSP, three refiners are designed to extract personalized tokens for response generation. Intuitively, the amount of the tokens will influence the refiner's performance. We report this influence in Figure~\ref{fig:his}.
As we can see in the left part, the quality of response generation improves with more tokens used. This is because fewer tokens are incapable of covering sufficient personalized information for response generation. Our MSP model performs optimally when about 200 personalized tokens are selected. When more tokens are introduced, the performance degrades. The potential reason is that more tokens would bring the noise to the generation. This is consistent with our speculation that the dialogue history is noisy and the information selection is both effective and necessary.

To provide a more qualitative view of our method, we conduct a case study in Appendix~\ref{sec:case}.

\section{Conclusion} \label{conclusion}
In this work, we propose an MSP model for personalized response generation.
Unlike previous related work, we utilize a refiner structure to extract query-aware persona information from large-scale dialogue history. The multi-level refiners can sparsely extract valuable information from dialogue history and leverage similar users' information to enhance the current user's personalization. Experimental results confirm the effectiveness of our model in generating informative and personalized responses.

\section*{Acknowledgement}
Zhicheng Dou is the corresponding author. This work was supported by National Natural Science Foundation of China No. 61872370 and No. 61832017,  Beijing Outstanding Young Scientist Program NO. BJJWZYJH012019100020098, and Intelligent Social Governance Platform, Major Innovation \& Planning Interdisciplinary Platform for the ``Double-First Class'' Initiative, Renmin University of China. We also acknowledge the support provided and contribution made by Public Policy and Decision-making Research Lab of RUC.


\clearpage
\appendix

\begin{table}[t]
\caption{The results of other evaluation metrics on Weibo and Reddit dataset. ``$\dagger$'' indicates that our model achieves significant improvement in t-test with $p$-value $<$ 0.05. The best results are in \textbf{bold}.}\label{tab: extra}
\small
\centering
\label{tab: extra}
\setlength{\tabcolsep}{0.7mm}{
\begin{tabular}{llrrrr}
\toprule
& & {BLEU-3} & {BLEU-4} & {ROUGE-1} & {ROUGE-2}\\ 
\midrule
\multirow{9}{*}{\rotatebox[origin=c]{90}{Weibo}} & Seq2Seq & 0.011$^\dagger$ & 0.001$^\dagger$ & 8.740$^\dagger$ & 0.373$^\dagger$ \\
 & MMI & 0.046$^\dagger$ & 0.007$^\dagger$ & 5.316$^\dagger$ & 0.105$^\dagger$ \\
 & DialoGPT & 0.114$^\dagger$ & 0.027$^\dagger$ & 9.414$^\dagger$  & 0.632$^\dagger$\\
 & GPMN & 0.359$^\dagger$ & 0.066$^\dagger$ & 8.086$^\dagger$ & 0.753$^\dagger$   \\
 & PerCVAE & 0.466$^\dagger$ & 0.089$^\dagger$& 7.946$^\dagger$ & 0.485$^\dagger$ \\
 & Speaker & 0.107$^\dagger$ & 0.041$^\dagger$ & 7.997$^\dagger$ & 0.155$^\dagger$\\
 & PersonaWAE & 0.889$^\dagger$ & 0.155$^\dagger$ & 11.341$^\dagger$ & 0.358$^\dagger$ \\
 & DHAP & 1.170$^\dagger$ & {0.401}$^\dagger$& 14.131$^\dagger$ & 3.608$^\dagger$ \\
 & \textbf{MSP} (Ours) & \textbf{3.973} & \textbf{3.522}& \textbf{16.249} & \textbf{5.812}  \\ 
\midrule
\multirow{10}{*}{\rotatebox[origin=c]{90}{Reddit}} & Seq2Seq & 0.007$^\dagger$ & 0.001$^\dagger$ & 3.989$^\dagger$  & 0.233$^\dagger$\\
 & MMI  & 0.007$^\dagger$& 0.003$^\dagger$ & 3.960$^\dagger$ & 0.245$^\dagger$ \\
 & DialoGPT & 0.054$^\dagger$ & 0.010$^\dagger$ & 8.977$^\dagger$ & 0.610$^\dagger$\\
 & GPMN & 0.039$^\dagger$ & 0.006$^\dagger$ & 4.896$^\dagger$ & 0.330$^\dagger$\\
 & PerCVAE & 0.068$^\dagger$ & 0.009$^\dagger$ & 8.004$^\dagger$ & 0.540$^\dagger$\\
 & Speaker & 0.021$^\dagger$ & 0.005$^\dagger$ & 4.017$^\dagger$ & 0.245$^\dagger$ \\
 & PersonaWAE & 0.029$^\dagger$ & 0.007$^\dagger$ & 8.247$^\dagger$ & 0.517$^\dagger$\\
 & DHAP & 0.079$^\dagger$ & 0.013$^\dagger$ & 10.680$^\dagger$ & 0.697$^\dagger$ \\
 & \textbf{MSP} (Ours) & \textbf{0.106} & \textbf{0.019} & \textbf{11.078} & \textbf{0.745}\\ 
\bottomrule
\end{tabular}%
}
\end{table}

\begin{table*}[t]
\centering
\small
\caption{A case study. Due to space limitation, we omit some user history responses and references.}
\label{tab: case study}
\begin{tabular}{lp{.8\linewidth}}
\toprule
\multirow{4}{*}{History} &\textbf{H1:} Liverpool's configuration has the life of a champion, support it! \\
 &\textbf{H2:} Champion Liverpool! \\
 & \textbf{H3:} I like to listen to music. \\
 &\textbf{H4:} The songs on this album are really beautiful, so worth enjoying.  \\ \midrule
Query  & New album "How Am I? -The sun rises as it always does", the first single "Ember" heals the system, go listen!  \\ \midrule
\multirow{2}{*}{Persona Reference}  & \textbf{R1:} I like to listen to music.  \\
&\textbf{R2:} The songs on this album are really beautiful, so worth enjoying.\\ \midrule
\multirow{2}{*}{Sim Reference}  & \textbf{R1:} Quite like this type of song.  \\ 
&\textbf{R2:} Angela Leung's "Ember" is really good. \\ \midrule
Persona Profile  & like, song, album, beautiful, enjoying  \\ \midrule
Sim Profile  & like, song, Angela Leung, good, Ember  \\  \midrule
\multirow{3}{*}{Response} & \textbf{DialoGPT}: My little heart has flown.  \\
 & \textbf{MSP(Ours)}: Angela Leung's song is very good. \\
 & \textbf{Golden}: Angela Leung's songs must be listened to.  \\ 
 \bottomrule
 \end{tabular}%
\end{table*}


\section{Additional Experimental Results}
\label{sec:extra}
As n-gram word overlap metrics can reflect user speaking style more accurately, we evaluate the BLEU-3/4~\citep{Salim_bleu_2002}, and the result is shown in Table~\ref{tab: extra}. It is consistent with other evaluations that our model outperforms every indicator. This demonstrates that user profiles also contain speaking style information, and our model can use the information to achieve a personalized response.


\section{Case Study}
\label{sec:case}
To show the effect of our model more concretely, we adopt a case study, and the results are shown in Table~\ref{tab: case study}. It shows that our model can extract profiles from both current and similar users and generate informative and personalized responses. Specifically, in his dialogue history, he mentioned sports $\textbf{H1, H2}$ and music $\textbf{H3, H4}$ topics. Firstly, we can select similar users who also talk about sports and music, using the user refiner. Then, as the query is related to the music topic, we extract the current and sim users' dialogue history responses about the music topic $\textbf{R1, R2}$ by the topic refiner. 
Furthermore, the token refiner selects some meaningful and personalized words from long sentences of reference. In this case, we can find that the token refiner extracts some compliment words (like, beautiful, enjoying) from the current user's history sentence since the current user likes listening to music. And the token refiner captures more concrete tokens from sim users' history sentences, such as ``Angela Leung'' and ``Ember''. By combing two profiles, our personalized generator gets an informative and personalized response close to ground-truth. In contrast, DialoGPT generates a fluent but meaningless response to the query.

\end{document}